\documentclass[journal,twoside,web,10pt]{ieeecolor}
\usepackage{generic}
\usepackage[mincitenames=1, maxcitenames=2, minbibnames=1, maxbibnames=2, backend=biber, style=ieee]{biblatex}
\usepackage{enumerate}
\usepackage[linesnumbered,ruled,vlined]{algorithm2e}
\usepackage{graphicx}
\usepackage{subfigure}
\usepackage{multirow, multicol}
\usepackage{booktabs}
\usepackage{amsmath, amssymb}
\newcommand{\mypara}[1]{
	\vspace*{0.01cm}
	\noindent\textbf{\textit{#1}}}

\DeclareSourcemap{
  \maps[datatype=bibtex]{
    \map[overwrite]{
      \step[fieldsource=month, match=\regexp{\A(j|J)an(uary)?\Z}, replace=1]
      \step[fieldsource=month, match=\regexp{\A(f|F)eb(ruary)?\Z}, replace=2]
      \step[fieldsource=month, match=\regexp{\A(m|M)ar(ch)?\Z}, replace=3]
      \step[fieldsource=month, match=\regexp{\A(a|A)pr(il)?\Z}, replace=4]
      \step[fieldsource=month, match=\regexp{\A(m|M)ay\Z}, replace=5]
      \step[fieldsource=month, match=\regexp{\A(j|J)un(e)?\Z}, replace=6]
      \step[fieldsource=month, match=\regexp{\A(j|J)ul(y)?\Z}, replace=7]
      \step[fieldsource=month, match=\regexp{\A(a|A)ug(ust)?\Z}, replace=8]
      \step[fieldsource=month, match=\regexp{\A(s|S)ep(tember)?\Z}, replace=9]
      \step[fieldsource=month, match=\regexp{\A(o|O)ct(ober)?\Z}, replace=10]
      \step[fieldsource=month, match=\regexp{\A(n|N)ov(ember)?\Z}, replace=11]
      \step[fieldsource=month, match=\regexp{\A(d|D)ec(ember)?\Z}, replace=12]
    }
  }
}

\begin{document}
\title{Deep Learning Based Walking Tasks Classification in Older Adults using fNIRS}
\author{
    Dongning~Ma,~\IEEEmembership{Student Member,~IEEE,}	
	Meltem~Izzetoglu,
	Roee~Holtzer,
	and~Xun~Jiao,~\IEEEmembership{Member,~IEEE,}
	\thanks{This paper is submitted for review on IEEE Transactions on Neural Systems and Rehabilitation Engineering.}
	\thanks{This work was supported in part by the National Institutes on Aging R01AG036921 and R01AG044007. D. Ma, M. Izzetoglu, and X. Jiao are with the Department of Electrical and Computer Engineering, Villanova University, Villanova, PA 19085 USA (e-mail: dma2@villanova.edu, mizzetog@villanova.edu xjiao@villanova.edu). R. Holtzer is with the Department of Neurology, Albert Einstein College of Medicine and Ferkauf Graduate School of Psychology of Yeshiva University, Bronx, NY 10461 USA, (e-mail: roee.holtzer@yu.edu)}
}
\maketitle

\begin{abstract}
Decline in gait features is common in older adults and an indicator of increased risk of disability, morbidity, and mortality. Under dual task walking (DTW) conditions, further degradation in the performance of both the gait and the secondary cognitive task were found in older adults which were significantly correlated to falls history. Cortical control of gait, specifically in the pre-frontal cortex (PFC) as measured by functional near infrared spectroscopy (fNIRS), during DTW in older adults has recently been studied. However, the automatic classification of differences in cognitive activations under single and dual task gait conditions has not been extensively studied yet. In this paper, we formulate this as a classification task and leverage deep learning to perform automatic classification of STW, DTW and single cognitive task (STA). We conduct analysis on the data samples which reveals the characteristics on the difference between HbO2 and Hb values that are subsequently used as additional features. We perform feature engineering to formulate the fNIRS features as a 3-channel image and apply various image processing techniques for data augmentation to enhance the performance of deep learning models. Experimental results show that pre-trained deep learning models that are fine-tuned using the collected fNIRS dataset together with gender and cognitive status information can achieve around 81\% classification accuracy which is about 10\% higher than the traditional machine learning algorithms. We further perform an ablation study to identify rankings of features such as the fNIRS levels and/or voxel locations on the contribution of the classification task. 
\end{abstract} 


\section{Introduction}
\IEEEPARstart{M}{obility} impairments are common older adults affecting their functional independence and leading to increased risk of disability, morbidity, and mortality ~\cite{studenski2011gait}. Studies have shown that attention which is sub-served by the Pre-Frontal Cortex (PFC) and its related circuits plays a key role in the higher order cognitive control of mobility~\cite{holtzer2012relationship, yogev-seligmann2008role, holtzer2014neuroimaging, amboni2013cognitive, lindenberger2000memorizing}. Especially under complex and more taxing locomotion tasks as in dual task walking (DTW), allocating attention to competing task demands, requires use of additional attentional resources~\cite{rohrer1998two}, can result in degradation in both the gait and the secondary task performances, and is sensitive to aging posing a key risk factor for incident frailty and falls~\cite{holtzer2004age, verghese2012mobility}. Hence, assessment and identification of cognitive resource allocation together with gait performance under simple and attention demanding walking conditions, can be critical for incident risk assessment and prevention of falls in normal aging as well as in disease populations.

Motor control models of locomotion and robust associations between structural changes in frontal and subcortical brain regions with mobility outcomes have been established~\cite{drew2004cortical, lucas2019moderating, wagshul2019multi-modal}. Even though converging evidence suggests the role cognitive processes, specifically the executive functions in explaining mobility performance and decline in older adults ~\cite{holtzer2012relationship, yogev-seligmann2008role}, studies on the real time assessment and specific detections of functional neural correlates of simple and attention-demanding locomotion tasks is scarce. This gap could be in part due to the requirements of subject immobility and supine positioning in traditional neuroimaging modalities during scanning procedures making functional imaging of real, on the ground walking unattainable.

Recent studies began to increasingly utilize an emerging neuroimaging modality, namely functional near infrared spectroscopy (fNIRS) to assess cortical control and functional correlates of mobility under simple and attention demanding dual task walking conditions in aging populations~\cite{holtzer2014neuroimaging, vitorio2017fnirs, herold2017functional, leff2011assessment, hamacher2015brain, holtzer2015online, holtzer2016neurological, holtzer2018distinct, holtzer2011fnirs, holtzer2017stress, holtzer2017interactions, chen2017neural, george2019effect, holtzer2018effect, hernandez2016brain, holtzer2020mild, izzetoglu2020effects}. fNIRS is an optics-based non-invasive, safe, portable, and wearable neuroimaging technique~\cite{bunce2006functional, izzetoglu2005functional, scholkmann2014review, cutini2012functional, quaresima2016functional}, which can monitor relative changes in oxygenated-hemoglobin (HbO2) and deoxygenated-hemoglobin (Hb) associated with cognitive activity in real world tasks such as walking and talking.

While the tasks used in the investigation of functional brain mechanisms of mobility using fNIRS technology varies across studies, the most commonly implemented ones involve balance tasks, running, climbing the stairs and STW and DTW conditions~\cite{holtzer2014neuroimaging, vitorio2017fnirs, herold2017functional, leff2011assessment, hamacher2015brain, holtzer2015online, holtzer2016neurological, holtzer2018distinct, holtzer2011fnirs, holtzer2017stress, holtzer2017interactions, chen2017neural, george2019effect, holtzer2018effect, hernandez2016brain, holtzer2020mild, izzetoglu2020effects}. Specifically, in prior fNIRS studies reproducible and statistically significant increases have been found in HbO2 obtained from the PFC in DTW as compared to STW due to greater cognitive demands on attentional resources and gait performance that are inherent in the DTW condition~\cite{holtzer2018distinct, holtzer2011fnirs, holtzer2017stress, holtzer2017interactions, chen2017neural, george2019effect, holtzer2018effect, hernandez2016brain, holtzer2020mild, izzetoglu2020effects}. Furthermore, it was found that cortical responses to task demands specifically in the DTW condition were moderated by age~\cite{holtzer2011fnirs}, gender and stress~\cite{holtzer2017stress}, fatigue level~\cite{holtzer2017interactions}, medication use~\cite{george2019effect}, and disease status including diabetes~\cite{holtzer2018effect}, Multiple Sclerosis (MS)~\cite{hernandez2016brain}, mild cognitive impairments~\cite{holtzer2020mild}, and neurological gait abnormalities~\cite{holtzer2016neurological}. 

Even though growing number of studies that utilized fNIRS measures on older adults have repeatedly shown that hemodynamic biomarkers from PFC can provide significant differences between STW and DTW conditions in healthy and disease populations, automatic classification of these tasks using machine learning algorithms have not yet been studied. Automatic detection  of attentionally more demanding vs simple walking tasks using discriminative hemodynamic features extracted from HbO2 and Hb can provide information on an individual’s use of his/her attentional resources during active walking. Such automated detection indicative of attentional load during active walking can help in real time identification or prediction of cognitive overload, loss of gait control, reduction in gait performance and even prevention of falls. Moreover, identification of selective features that can discriminate walking task conditions can also lead to further diagnosis, monitoring and automatic classification of different age-related disease conditions where PFC activations in DTW were found to differ.

fNIRS measures have been used in the classification of wide range of tasks and disease populations in different age groups in prior studies. Some of these applications involve monitoring of mental workload, motor imagery, auditory and visual perception, various brain computer interfaces, pain assessment, anesthesia monitoring, attention deficit and hyperactivity disorder (ADHD) diagnosis, cognitive decline in traumatic brain injury, diagnosis of various mental illnesses such as schizophrenia~\cite{liu2017multisubject, putze2014hybrid, chiarelli2018deep, naseer2015fnirs-based, pourshoghi2016application, hernandez-meza2018investigation, guven2019combining, merzagora2019functional, song2017automatic}. However, there are very few studies on the classification of gait related tasks. Existing studies primarily monitored motor areas and investigated classification of intention or preparation to different types of gait in healthy young adults primarily for gait rehabilitation applications involving control of assistive devices where classification accuracy was found in about 80\% ranges~\cite{jin2018pilot, li2020detecting, rea2014lower, khan2018fnirs-based}. In these small number of prior studies, fNIRS measures from PFC during single and attentionally demanding dual task active walking conditions that are indicative of different attentional states and cognitive load conditions in elderly populations were not studied with machine learning models. 

In this study our aim is to achieve automatic classification of walking tasks requiring different levels of cognitive resources. Specifically, we develop a comprehensive pipeline for processing and engineering the collected fNIRS data to efficiently extract the features. We fine-tune pre-trained state-of-the-art deep learning models over the fNIRS dataset and obtain up to 81\% accuracy, which is about 10\% higher than the traditional machine learning algorithms. We also conduct ablation studies for identifying critical features when using fNIRS for classifying walking tasks of older adults.

This paper is organized as follows: In Section~\ref{sec:pt}, we introduce the information of the participants and our task protocol. In Section~\ref{sec:m}, we explain our proposed methods in detail. We present the results of our comprehensive results in Section~\ref{sec:r} and finally, we provide concluding remarks in Section~\ref{sec:c}. To the best of our knowledge, we are the first to apply deep learning methods in fNIRS-based walking task classification for older adults. 

\section{Participants and Task Protocol}
\label{sec:pt}
\subsection{Participants}
The study involved a total of $n=451$ community dwelling older adults in Lower Westchester county, NY of age 65 years and older (76.16 $\pm$ 6.67, 223 females) who were originally enrolled in a longitudinal cohort study entitled ``Central Control of Mobility in Aging'' (CCMA)~\cite{holtzer2014neuroimaging, holtzer2015online}. Recruitment procedures started with the identification of potential participants from population lists and then conducting a structured telephone interview to obtain verbal assent, assess medical history, mobility and cognitive functioning. Participants with significant loss of vision and/or hearing, inability to ambulate independently, current or history of severe neurological or psychiatric disorders, and recent or anticipated medical procedures that may affect mobility were excluded from the study. Individuals who agreed to participate in the study, fell into the inclusion/exclusion criteria and passed the phone interview were invited to two annual in-person study visits each lasting around 3 hours at the research center at Albert Einstein College of Medicine, Bronx, NY. During these visits, participants received a structured neurological examination and comprehensive neuropsychological, psychological, functional, and mobility assessments. Functional brain monitoring using fNIRS during the STW and DTW protocol was completed in one session. Cognitive status was determined at consensus diagnostic case conferences~\cite{holtzer2008within-person}. Repeatable Battery for the Assessment of Neuropsychological Status (RBANS) was used to characterize overall level of cognitive function~\cite{duff2008utility}. The sample was relatively healthy (Global Health Status mean score = 1.62 $\pm$ 1.09) and in the average range of overall cognitive function (RBANS mean Index score = 91.77 $\pm$ 11.71). The work described in this manuscript has been executed in adherence with The Code of Ethics of the World Medical Association (Declaration of Helsinki) and the APA ethical standards set for research involving human participants. Written informed consents were obtained at the first clinic visit according to study protocols approved by the Institutional Review Board at Albert Einstein College of Medicine, Bronx, NY (Protocol \#2010-224; Date: 03/03/2022). 

\subsection{Task Protocol}
The task protocol used in this study involved two single tasks and one dual-task conditions presented in a counterbalanced order using a Latin-square design to minimize task order effects on the outcome measures. The single task conditions were 1) single-task walking (STW) and 2) the single task alpha cognitive interference task (STA). In STW condition, participants were asked to walk at their “normal pace” around a 4 $\times$ 20 foot electronic walkway (Zenometrics system with Zeno electronic walkway using ProtoKinetics Movement Analysis Software (PKMAS), Zenometrics, LLC; Peekskill, NY). In the Alpha condition participants were asked to stand still on the electronic walkway while reciting alternate letters of the alphabet out loud (A, C, E…) for 30 seconds. In the dual-task walking (DTW) condition, participants were required to perform the two single tasks at the same time by walking around the walkway at their normal pace while reciting alternate letters of the alphabet. Participants were specifically asked to pay equal attention to both the walking and cognitive interference tasks to minimize task prioritization effects. In both STW and DTW conditions participants were asked to walk on the instrumented walkway in three continuous loops that consisted of six straight walks and five left-sided turns. The duration of each task condition varied depending on the individual’s walking speed. Reliability and validity for this walking paradigm have been well established~\cite{holtzer2014performance}.

\section{Methods}
\label{sec:m}
\begin{figure*}[htbp]
    \centering
    \subfigure{\includegraphics[keepaspectratio, width=2\columnwidth, page=1]{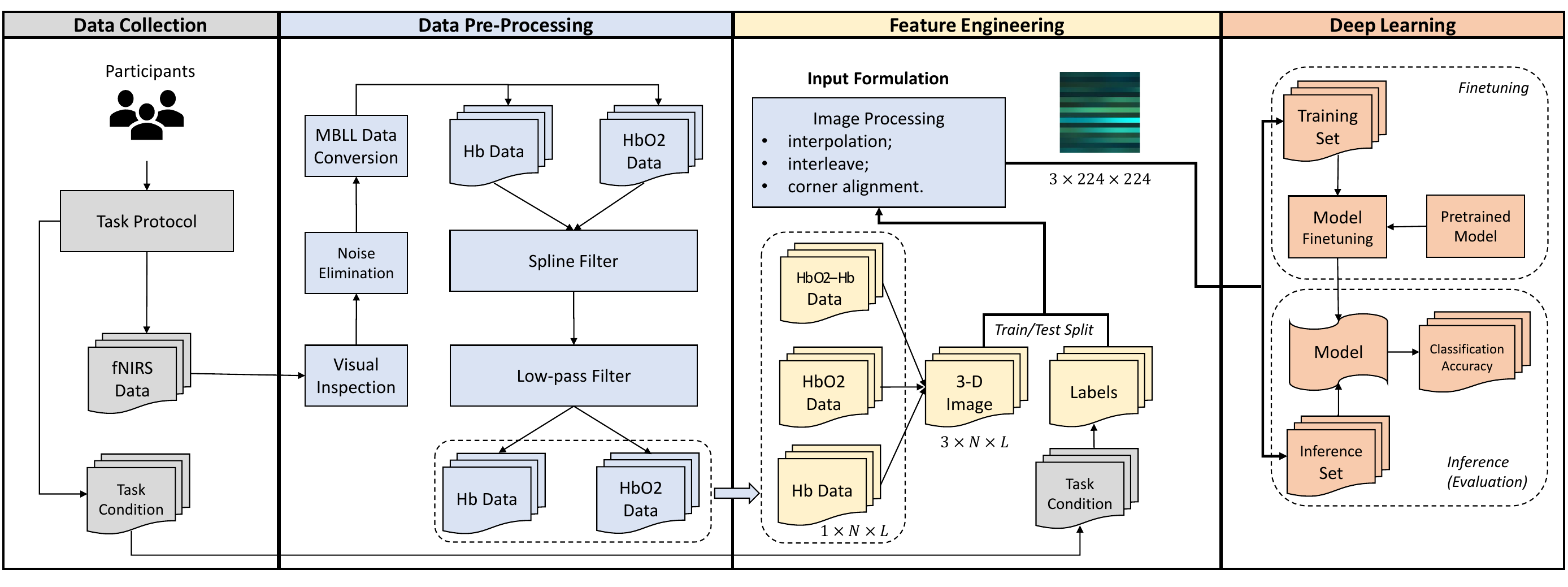}}
    \caption{Overview of the proposed framework. There are 4 major phases: 1) \textbf{Data Collection}: Subjects are asked to perform task protocols and fNIRS data and task condition labels are collected. 2). \textbf{Data Pre-processing}: Signal processing methods are applied for de-noising, conversion and filtering on the raw collected fNIRS data. 3). \textbf{Feature Engineering}: fNIRS data are formulated and engineered into an image which is ready for deep learning model to learn the features. 4). \textbf{Deep Learning}: Various deep learning models are trained or fine-tuned on the fNIRS dataset. The models are evaluated on a separate inference set on their task condition classification accuracy.}
    \label{fig:ov}
\end{figure*}

An overview of the proposed methods utilized in this work is illustrated in Fig.~\ref{fig:ov}. We have four major steps: data collection, data pre-processing, feature extraction and deep learning:
\begin{itemize}
    \item \textbf{Data Collection:} In data collection, participants were asked to complete the task protocol as instructed, during which their hemodynamic activations were collected using fNIRS. In addition, we also collected subject-related data (gender and RBANS).
    \item \textbf{Data Pre-processing:} In data pre-processing, we applied different methods such as visual inspection, wavelet denoising, hemodynamic data conversion, and spline and low pass filterings to obtain HbO2 and Hb data of participants in time domain for different task conditions.
    \item \textbf{Feature Engineering:} We formulated the data pre-processed as an image tensor. First two channels of the image tensor represent the pre-processed Hb and HbO2 data, respectively. We also added another channel into the image tensor which is the difference between the HbO2 and Hb, i.e, the HbO2 - Hb referred to as the oxigen index~\cite{izzetoglu2005motion}.
    \item \textbf{Deep Learning:} We applied various deep learning algorithms using the PyTorch framework~\cite{paszke2019pytorch}. We used the pre-trained deep neural network and vision transformer architectures which are available open-source, and fine-tune them with the engineered fNIRS data and evaluate the model performance.
\end{itemize}

\subsection{Data Collection}
\begin{figure}[htbp]
    \centering
        \subfigure{\includegraphics[keepaspectratio, width=.52\columnwidth, page=1]{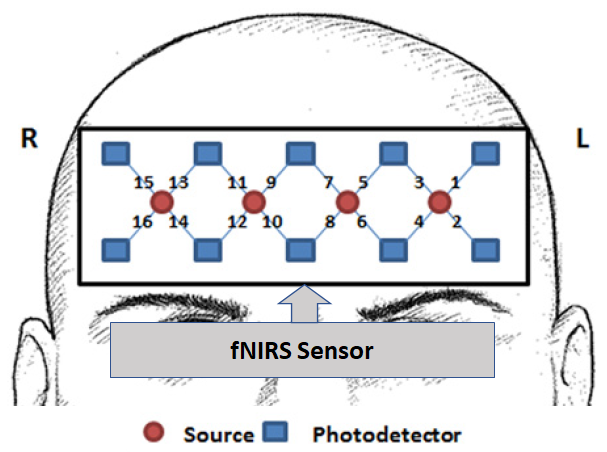}}
    \caption{fNIRS system of the sensor pad and the sensor placement on the forehead with 16 voxel locations.}
    \label{fig:fnirs_sys}
\end{figure}

\mypara{fNIRS System.} We have utilized the fNIRS Imager 1100 (fNIRS Devices, LLC, Potomac, MD) in this study to collect the hemodynamic activations in the PFC while participants were performing the task protocol~\cite{holtzer2015online, bunce2006functional, izzetoglu2005functional, ayaz2011using}. In this fNIRS device, the sensor consists of 4 LED light sources and 10 photodetectors configured as shown in Fig.~\ref{fig:fnirs_sys} where each source-detector separation is set to 2.5 cm. The light sources on the sensor (Epitex Inc. type L4X730/4X805/4X850-40Q96-I) contain three built-in LEDs having peak wavelengths at 730, 805, and 850 nm, with an overall outer diameter of 9.2 $\pm$ 0.2 mm. The photodetectors (Texas Instruments, Inc., type OPT101) are monolithic photodiodes with a single supply transimpedance amplifier. With the given source-detector configuration and the serial data collection regime of the device, hemodynamic changes in the PFC can be monitored at the sampling rate of 2 Hz with 16 voxels as shown in Fig.~\ref{fig:fnirs_sys}. 

During the fNIRS data collection procedure, first the fNIRS sensor was placed on the forehead of the recruited participants. A standardized sensor placement procedure based on landmarks from the international 10–20 system was implemented~\cite{ayaz2011using, ayaz2006registering} where middle of the sensor was aligned with the nose horizontally and the bottom of the sensor was placed above the eyebrows vertically. Testing was conducted in a quiet room. Participants wore comfortable footwear and performed the task protocol with the fNIRS sensor attached to their forehead during the overall data collection period.

\subsection{Data Pre-processing}
First, visual inspection was performed on individual data from all voxels to identify and eliminate the ones with saturation, dark current conditions or extreme noise. Then to eliminate spiky type noise, wavelet denoising with Daubechies 5 (db5) wavelet was applied to the raw intensity measurements at 730 and 850 nm wavelengths as proposed in~\cite{molavi2012wavelet-based} and widely applied in fNIRS studies~\cite{chiarelli2015kurtosis-based}. The artifact-removed raw intensity measurements were then converted to changes in HbO2 and Hb using modified Beer-Lambert law (MBLL)~\cite{izzetoglu2020effects, izzetoglu2005functional, cope1988system}. In MBLL, previously published values for conversion parameters i.e. wavelength and chromophore dependent molar extinction coefficients ($\epsilon$) and age and wavelength adjusted differential pathlength factor (DPF) were used~\cite{izzetoglu2020effects, izzetoglu2005functional, scholkmann2013general}. Finally, we applied Spline filtering~\cite{scholkmann2010how} followed by a finite impulse response low-pass filter with cut-off frequency at 0.08 Hz~\cite{izzetoglu2020effects, yucel2016mayer} to HbO2 and Hb data separately to remove possible baseline shifts and to suppress physiological artifacts such as respiration and Mayer waves. 

Data epochs corresponding to each task condition, STW, STA and DTW, were extracted to be used in further processing for feature extraction and machine learning model generation for automatic activity classification. fNIRS data acquisition and the electronic walkway system for gait analysis were synchronized using a central ``hub'' computer with E-Prime 2.0 software where time stamps of start and end points for each baseline and task condition were marked and recorded~\cite{holtzer2018distinct, holtzer2011fnirs, holtzer2017stress, holtzer2017interactions, chen2017neural, george2019effect, holtzer2018effect, hernandez2016brain, holtzer2020mild, izzetoglu2020effects}. In order to correctly extract the data epochs during the exact walking task execution periods, a second level processing time synchronization method was implemented. The HbO2 and Hb data epochs corresponding to time interval between the first recorded foot contact with the walkway until the end of the 6th and final straight walk algorithmically determined by PKMAS as previously described in~\cite{holtzer2016neurological} were extracted for STW and DTW conditions. 
Finally, proximal 10-second baselines administered prior to each experimental task were used to determine the relative task-related changes in the extracted HbO2 and Hb data epochs for each of the task condition using the previously described baseline correction method (subtracting the average value of the proximal baseline region data from the following task epoch data)~\cite{holtzer2018distinct, holtzer2011fnirs, holtzer2017stress, holtzer2017interactions, chen2017neural, george2019effect, holtzer2018effect, hernandez2016brain, holtzer2020mild, izzetoglu2020effects}. We then used HbO2 and Hb data epochs in DTW, STW and STA tasks in further feature extraction and machine learning model development to automatically classify these three tasks in this work.

\subsection{Feature Engineering}
In this sub-section, we first show the statistical analysis across and within subjects, then present the feature engineering including input formulation for subsequent deep learning algorithms. 

\begin{figure}[htbp]
    \centering
    \includegraphics[keepaspectratio, width=.94\columnwidth, page=1]{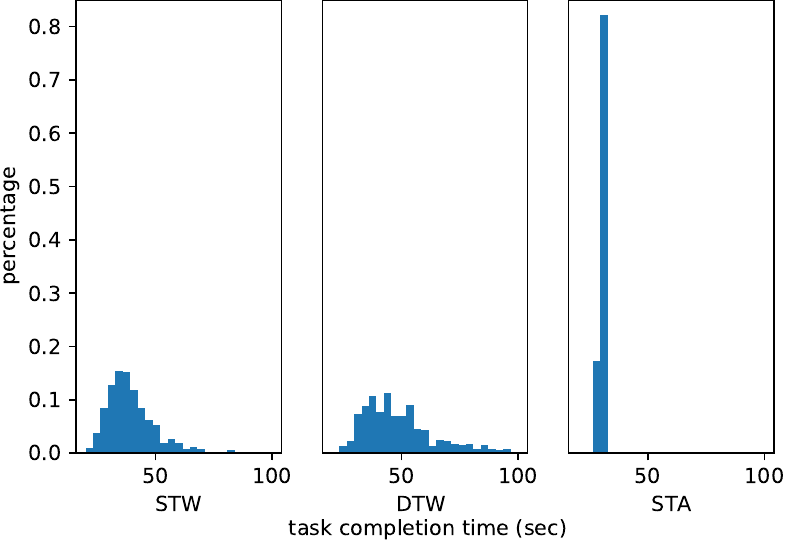}
    \caption{Histogram of task completion time of different subjects under three conditions.}
    \label{fig:completion}
\end{figure}

\mypara{Analysis across Subjects.}
We first illustrate the distribution of task completion time across subjects in Fig.~\ref{fig:completion}. The task completion time differs between subjects and task conditions due to individual variability and normal pace. Note that completion time of the STA is 30 seconds as defined in the task protocol. For the other two tasks, we notice that DTW on average requires more time to complete than STW, since DTW could be a more challenging task for older adults.

\begin{figure}[htbp]
    \centering
    \subfigure[Hb]{
        \includegraphics[keepaspectratio, width=.92\columnwidth, page=1]{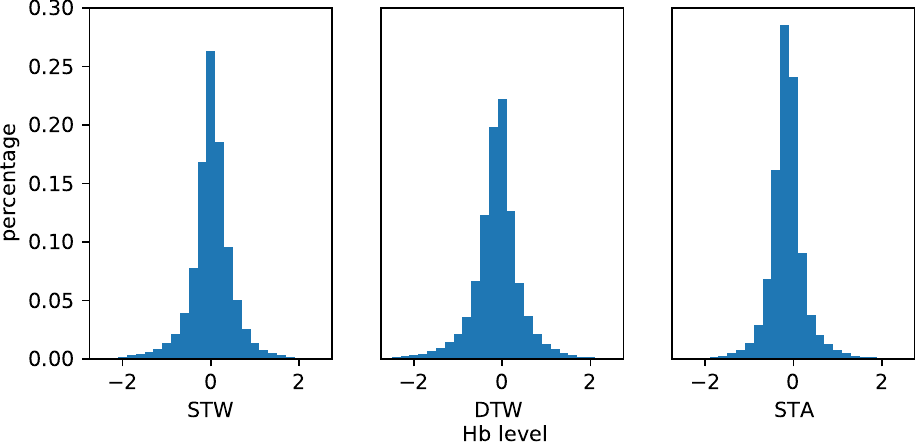}
    }
    \subfigure[HbO2]{
        \includegraphics[keepaspectratio, width=.92\columnwidth, page=1]{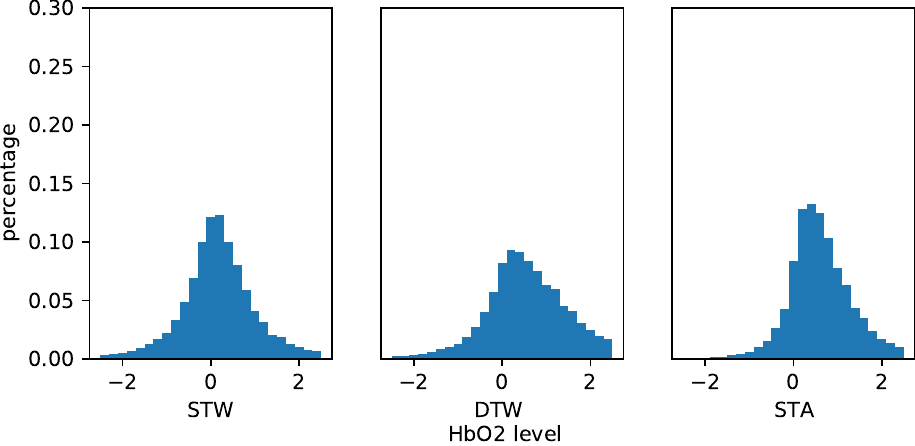}
    }
    \caption{Histogram of Hb and HbO2 levels of different subjects under three task conditions.}
    \label{fig:val}
\end{figure}

Additionally, we investigate the distribution of Hb and HbO2 values under the three task conditions as provided in the histogram plots in Fig.~\ref{fig:val}. For Hb levels, all the three task conditions exhibit a Gaussian-like distribution where the averages are around negative values close to 0, suggesting decreases relative to baseline conditions. DTW condition shows higher standard deviation than the rest of the two single tasks, suggesting more individual variability in this condition. For HbO2, distribution of STW is still showing Gaussian-like pattern with average around the positive values close to 0. However, for DTW and STA, the distributions have shifted to more positive values suggesting more cognitive activations relative to baseline in these task conditions as compared to STW. Comparatively, levels in DTW show larger shift than STA. Such distribution differences between Hb, HbO2 and HbO2 - Hb levels inspire us in this study to use them as additional features for the machine learning model.

\begin{figure*}[htbp]
    \centering
    \includegraphics[keepaspectratio, width=1.95\columnwidth, page=1]{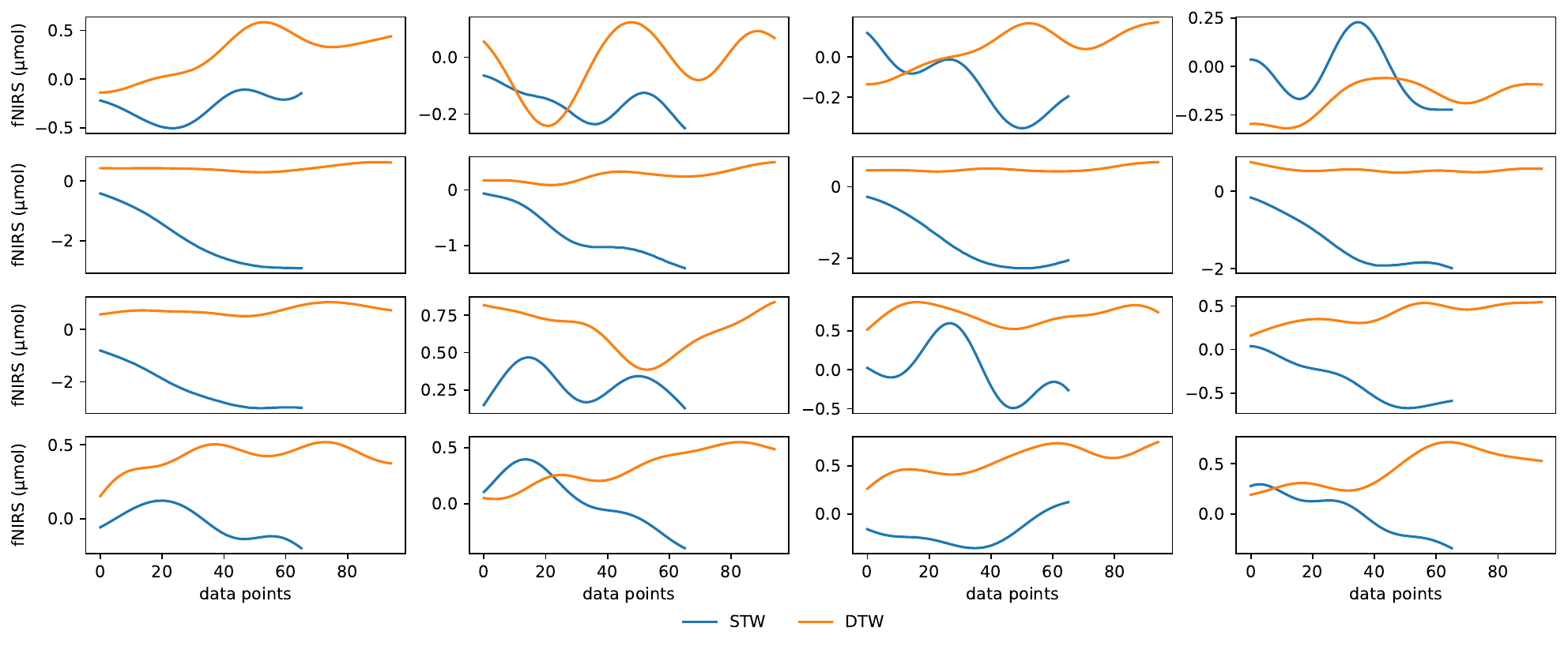}
    \caption{Representative case analysis: fNIRS HbO2 levels from 16 voxel locations of a randomly selected subject.}
    \label{fig:representative_o2}
\end{figure*}

\mypara{Analysis within Subject.}
Although Fig.~\ref{fig:val} indicates that the difference in hemodynamic activity levels of STW and DTW conditions can be leveraged as features, they could be different at the granularity of each voxel location for each individual subject. A representative subject HbO2 recording under STW and DTW conditions based on each voxel location is shown in Fig.~\ref{fig:representative_o2}. Note that, for this case, the time to complete DTW is higher than STW. We can observe from the plots that, although HbO2 levels in DTW are in general higher than those of STW, for each voxel location the range of values can drastically vary. In some channels and sample points, it can also be observed that HbO2 levels in STW are similar or even higher than DTW. Hence, there is a need to leverage the power of deep learning models beyond mere visual or statistical analysis to enable more accurate and individualized study, particularly to leverage the rich information revealed with each voxel location.

\mypara{Input Formulation.}
Based on the observation from the analysis, we aggregate the HbO2, Hb and HbO2 - Hb levels together and formulate each sample as a 3-channel image as input to the subsequent machine learning model. The first and the second channel are the HbO2 and Hb levels respectively and the third channel is the difference between HbO2 and Hb levels. For $i$-th sample in the dataset of each subject and task condition, the size is $3 \times N_i \times L_i$. $N_i \leq 16$ refers to the number of forehead voxel locations and $L_i = 2 \times T_i$ refers to the sample points which is twice the task completion time since we use sampling frequency of 2 Hz.

Although there are 16 voxel locations, due to limitations on data collection with excessive artifacts that cannot be cleaned, there are missing fNIRS data of one or a few voxel locations in some samples. We perform a screening and accept samples with no more than 3 missing voxel locations and any samples with missing numbers beyond 3 is regarded as invalid data and thus not used in training or inference. The number of valid samples in the dataset after removing is $n = 1216$. For the missing data we use interpolation to reconstruct the data into an image of size $3 \times 16 \times L_i$. 

Since the dataset is relatively small compared with general computer vision datasets with usually more than tens of thousands of images, we use pre-trained deep learning models rather than training models afresh to prevent over-fiting. Additionally, due to the difference on task completion time, the dimensions of the images are not consistent. Therefore, we perform further image processing to conform the input image dimensions with the input of the deep learning models that are usually pre-trained with the ImageNet dataset~\cite{deng2009imagenet}, i.e., $3 \times 224 \times 224$. Although such processing or resizing of inputs are common in deep learning, how the images are resized into $3 \times 224 \times 224$ can potentially have impact on the learning performance~\cite{talebi2021learning}. Note that, for each sample, data beyond first 224 points are truncated while data with less than 224 points are padded with 0. We intentionally avoid row-wise interpolation because we want the fNIRS data to keep the information of task completion time and pace of each subject which can be an important marker.

\begin{figure*}[htbp]
    \centering
    \subfigure[bilinear, aligned]{
        \includegraphics[keepaspectratio, width=.31\columnwidth, page=1]{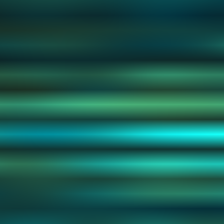}
    }
    \subfigure[bilinear, not aligned]{
        \includegraphics[keepaspectratio, width=.31\columnwidth, page=1]{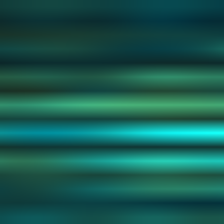}
    }
    \subfigure[bicubic, aligned]{
    \includegraphics[keepaspectratio, width=.31\columnwidth, page=1]{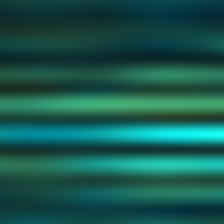}
    }
    \subfigure[bicubic, not aligned]{
        \includegraphics[keepaspectratio, width=.31\columnwidth, page=1]{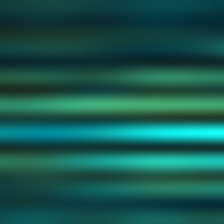}
    }
    \subfigure[interleave]{
        \includegraphics[keepaspectratio, width=.31\columnwidth, page=1]{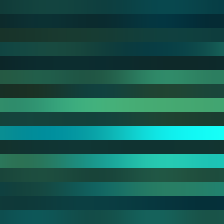}
    }
    \caption{Visualization of different image processing effects. Colors reflect the fNIRS values, i.e., Hb, HbO2 and HbO2 - Hb levels in each dimension.}
        \label{fig:augmentation}
\end{figure*}

In order to find the optimal processing configuration, we sweep across 3 different parameters which are two commonly used resizing techniques for deep learning model inputs~\cite{talebi2021learning}, which enables us 5 different configurations. The 3 parameters are visualized in Fig.~\ref{fig:augmentation} and also explained below:
\begin{itemize}
    \item \textbf{Interpolation mode}: Interpolation is to insert data between each original data to enlarge the input size into $3 \times 224 \times 224$. The available options are \textbf{bilinear} or \textbf{bicubic} interpolation.
    \item \textbf{Corner alignment}: This parameter decides whether to force corner alignment during interpolation. If true, the levels from the first and last voxel locations will be used as corner so that no interpolated values will appear as corners of the processed image. 
    \item \textbf{Interleave repeat}:  As we have 16 voxel locations, we are able to interleave repeat the data of each location by 14 times to achieve the required 224 dimension (row-wise) as an alternative for interpolation. Interleave repeat will duplicate exactly the data row-wise, which will reflect as the clear boarders between each voxel locations visually as illustrated in Fig.~\ref{fig:augmentation}. 

\end{itemize}

In summary, after feature engineering, we are able process and enhance the original samples with fNIRS levels of missing data and inconsistent dimensions into images in the shape of $3 \times 224 \times 224$, which is ready for the deep learning models. 

\subsection{Deep Learning}
We leverage the state-of-the-art deep learning models for machine vision and/or image classification. We first train the models afresh, i.e., without any pre-training. However, since the scale of the fNIRS dataset is rather small, the model easily overfits: the accuracy on training set achieves more than 90\% while the inference set is only around 70\%.

This is a common issue for other applications from medical imaging in the bioinformatics domain as the clinical data are usually much less than large general image datasets. Thus, for domain specific application with proprietary datasets, the concept of transfer learning~\cite{zhuang2020comprehensive} is usually in favor: the model is first pre-trained using the general and public image dataset and then fine-tuned on the proprietary dataset. In this work, using ResNet as an example, we obtain open-sourced, publicly available ResNet model which is trained using the ImageNet dataset which has 1000 classes. We then change the dimensionality of the last classifier (the fully connected layer) from 1000 to 3 to match the three task conditions. We fine-tune over this model by using the processed images to fit the model on the fNIRS dataset for classifying the three task conditions.

We implement our deep learning model using various architectures that are basically in two families: deep convolutional neural networks and vision transformer attention networks. Model details and configurations are introduced in the experimental results section Sec.~\ref{sec:r}.

\section{Experimental Results}
\label{sec:r}
In this section, we present the experimental results on the machine learning model for the classification of tasks in older adults. We evaluate the impact of different feature combinations as well as different machine learning algorithms on classification accuracy and computational efficiency.

\subsection{Experimental Setup}
\mypara{Environment.} We use Pytorch, a machine learning framework for python~\cite{paszke2019pytorch} to implement the machine learning models. We split the samples into training and inference set by 8:2, and the results are obtained with 5-fold cross validation. We use Adam optimizer with 0.001 learning rate and halt fine-tuning when accuracy does not increase further, thus the total number of epochs as well as the time for fine-tuning can vary across different models. 

\mypara{Models.}
We use the state-of-the-art deep learning and/or machine vision models from two architectures: deep neural networks and vision transformer attention networks. We sweep across different architectures including deep convolutional neural networks such as ResNet~\cite{he2016deep}, VGG~\cite{simonyan2014very}, MobileNet~\cite{sandler2018mobilenetv2}, EfficientNet~\cite{tan2019efficientnet}, and TinyNet~\cite{han2020model}. We also evaluate on recently emerging attention based vision transformer models~\cite{dosovitskiy2020image}. All the pretrained deep learning models are fetched from online open source repositories via PyTorch Image Models (timm) package~\cite{rw2019timm}. 

As a comparative study, we also implement several baseline machine learning models, including traditional machine learning models of decision tree, random forest (with 25 trees), and k-nearest neighbors (k = 5) which are implemented via Scikit-Learn package~\cite{pedregosa2011scikit}.

\subsection{Classification Results}
\mypara{Comparison with Baselines.} Results of this classification task including accuracy and error bars with different learning models (baseline and deep learning) is presented in Table~\ref{tab:acc}. 

We can first observe that deep learning models are able to out-perform all the baseline traditional learning algorithms by at least 10\% on the inference set. Particularly, decision tree and random forest are extremely over-fit on training data with higher than 99\% accuracy while having poor performance on inference set. K-nearest neighbors can hardly learn efficiently as the accuracy on training set is only around 77\% which is quite lower than all the other models.

Some of the deep learning models also experience different degree of overfit. Particularly for larger models such as VGG, ResNet and ViT-Base, their accuracy on training set is mostly over 94\% except for VGG. Smaller models like MobileNetV2, TinyNet and EfficientNet are usually less overfit as their accuracy is 90\% -- 93\% on training set. 

The top three models according to accuracy on inference set are: ResNet-18, VGG-13 and TinyNet-E. All the three models can achieve accuracy near 81\%. We use these three models for the following ablation study and efficiency analysis.

\begin{table}[htbp]
  \centering
  \caption{Comparison between Models of on Train (Fine-tune) and Inference Accuracy}
    \begin{tabular}{ccc}
    \toprule
    {Models} & {Train Acc.} & {Infer. Acc.} \\
    \midrule
    Decision Tree & 0.994$\pm$0.002 & 0.628$\pm$0.01 \\
    Random Forest (n\_tree = 25) & 0.997$\pm$0.001 & 0.662$\pm$0.011 \\
    k-Nearest Neighbors (k=5) & 0.767$\pm$0.028 & 0.646$\pm$0.004 \\
    \midrule
    ResNet-18 & 0.940$\pm$0.019 & \boldmath{}\textbf{0.810$\pm$0.023}\unboldmath{} \\
    ResNet-26 & 0.944$\pm$0.018 & 0.800$\pm$0.016 \\
    VGG-13 & 0.905$\pm$0.022 & \boldmath{}\textbf{0.808$\pm$0.022}\unboldmath{} \\
    VGG-16 & 0.899$\pm$0.008 & 0.800$\pm$0.006 \\
    ViT-Base & 0.962$\pm$0.007 & 0.794$\pm$0.017 \\
    \midrule
    TinyNet-E & 0.906$\pm$0.012 & \boldmath{}\textbf{0.810$\pm$0.017}\unboldmath{} \\
    MobileNetV2-050 & 0.923$\pm$0.015 & 0.791$\pm$0.024 \\
    EfficientNet-b0 & 0.927$\pm$0.036 & 0.751$\pm$0.106 \\
    ViT-Tiny & 0.968$\pm$0.009 & 0.796$\pm$0.013 \\
    \bottomrule
    \end{tabular}%
  \label{tab:acc}%
\end{table}%

\subsection{Ablation Study}
\begin{figure}[htbp]
    % \centering
    \includegraphics[keepaspectratio, width=.85\columnwidth, page=1]{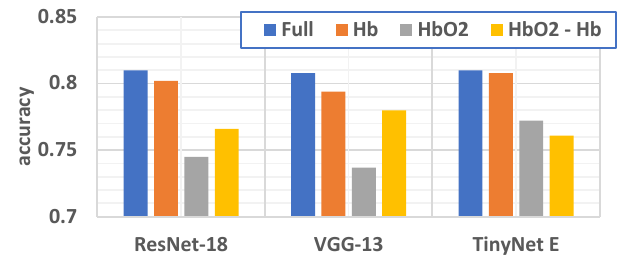}
    \caption{Accuracy by \textbf{removing} data from specific dimension separately. ``Full'' refers to no removal of any data. ``Hb'', ``HbO2'' and ``HbO2 - Hb'' refer to removal of ``Hb'', ``HbO2'' and ``HbO2 - Hb'' features from the overal analysis, respectively. }
    \label{fig:channel}
\end{figure}

\mypara{Dimension.}
We perform an ablation study to identify the contribution of features. First, we are to rank the contribution of each dimension in the input sample, e.g., Hb, HbO2 and the difference between HbO2 and Hb, respectively, to the classification performance. Therefore, we attempt to remove one dimension in the image while keeping the data in other two dimensions intact and observe the change on the accuracy of inference set. According to Fig.~\ref{fig:channel}, removing any of the three dimensions will cause accuracy degradation. However, model accuracy has different sensitivity towards different dimensions. Specifically, removing the HbO2 and the HbO2 - Hb dimension induces larger accuracy drop than removing Hb. This aligns with the statistical analysis in Fig.~\ref{fig:val} that HbO2 levels are different in distribution which indicate more discriminative features while Hb levels preserves similar Gaussian distribution under three task conditions and hence not very selective. Our results are also in line with the prior findings that HbO2 and hence the oxygenation are more reliable and sensitive to locomotion-related changes in cerebral blood flow~\cite{harada2009gait} and therefore providing the most distinctive features. 

\begin{table*}[htbp]
  \centering
  \caption{Task classification accuracy comparison under different image processing configurations}
    \begin{tabular}{cccccc}
    \toprule
    Models & bilinear, aligned & bilinear, not aligned & bicubic, aligned & bicubic, not aligned & interleave \\
    \midrule
    ResNet-18 & \textbf{0.81} & 0.8   & 0.808 & 0.797 & 0.802 \\
    VGG-13 & 0.806 & \textbf{0.808} & 0.798 & 0.804 & 0.798 \\
    TinyNet E & 0.799 & 0.795 & \textbf{0.81} & 0.806 & 0.802 \\
    \bottomrule
    \end{tabular}%
  \label{tab:abl_aug}%
\end{table*}%

\mypara{Image Processing.}
As shown in Fig.~\ref{fig:augmentation} (images are normalized for visualization purposes), we apply in total 5 different configurations on image processing. As an ablation study, we analyze the performance impact of image processing by comparing the inference set accuracy under different configurations in Table~\ref{tab:abl_aug}. We can observe that although the accuracy can vary up to 2\% across different configurations, we do not observe any single configuration able to dominate over other configurations. Based on such observation, we conclude that for different deep learning models, the best configuration can vary and require individual evaluation to select for the best to extract the features from the fNIRS data.

\begin{figure}[htbp]
    \centering
    \includegraphics[keepaspectratio, width=.76\columnwidth, page=1]{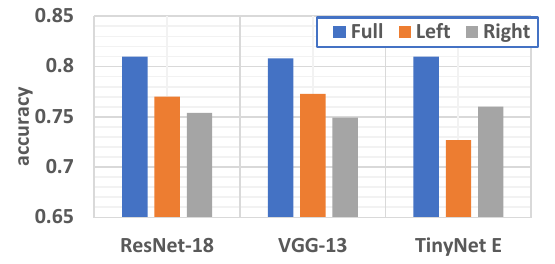}
    \caption{Task classification accuracy by \textbf{removing} data from specific hemisphere. ``Full'' refers to no removal of any data. ``Left'' and ``Right'' refer to removal of features from ``Left'' and ``Right'' hemisphere from the overall analysis, respectively. }
    \label{fig:hemi}
\end{figure}

\mypara{Hemisphere.}
We also try to characterize hemispheric contributions to the classification accuracy. We prune the input samples by removing all the data from voxel locations that is from one hemisphere (Channel 1 - 8 for left and 9 - 16 for right) and only use the rest for deep learning and identify the model performance. Based on Fig.~\ref{fig:hemi}, we can observe that for TinyNet-E model, data from the left hemisphere seem to contribute more while for ResNet-18 and VGG-13 models, removing data from right hemisphere causes more accuracy drop. In general, by removing data from either hemisphere will result in accuracy degradation for up to 8\%, thus indicating it is preferred to use the data from all the voxel locations for better model performance.

\subsection{Model Efficiency}
We also provide insights on the three models for their cost and overhead including number of parameters, model size, time for model fine-tuning and inference as well as the throughput (samples per second) in Table~\ref{tab:efficiency}. Time and throughput data are obtained with NVIDIA Tesla P100 GPU. 

ResNet-18 and VGG-13 use relatively larger model size and show lower throughput. Particularly for VGG-13, although the model is quite large with around 130M parameters, it does not show more competitive accuracy than the rest. For TinyNet-E, since the classifier layer (fully connected layer) output is reduced to 3, the model size becomes drastically compact and the throughput nearly triples the rest two models. However, based on our experiment, TinyNet-E requires more epochs to achieve comparable accuracy, thus even if it is smaller in model, it takes more time than ResNet-18 to fine-tune. 

\begin{table}[htbp]
  \centering
  \caption{Efficiency Comparison of Deep Learning Models}
    \begin{tabular}{ccccc}
    \toprule
    \multirow{2}[2]{*}{Models} & \multirow{2}[2]{*}{Params} & Model Size & Finetune/Infer & Throughput \\
          &       & (MB)  &  Time (sec) & (sample/sec) \\
    \midrule
    ResNet-18 & 11.18M & 42.7  & 55.1/0.250 & 4.9K \\
    VGG-13 & 129M  & 491.9 & 180.5/0.459 & 2.6K \\
    TinyNet E & 0.8M  & 2.9   & 78.4/0.089 & 13.7K \\
    \bottomrule
    \end{tabular}%
    
  \label{tab:efficiency}%
\end{table}%

\section{Conclusion}
Functional near infrared spectroscopy (fNIRS) is an optic-based, non-invasive neuroimaging modality, which is increasingly used as a safe and portable method to assess the cortical control of gait. Notably, fNIRS studies repeatedly show increased activation in the prefrontal cortex from single task walk (STW) to dual-task walk (DTW) conditions in older adults due to increased attentional demands in DTW, which is also an established risk factor for incident frailty, disability, and mortality. In this paper, we introduce and integrate the emerging deep learning methods into the pipeline of using fNIRS measures based on oxygenated (HbO2) and deoxygenated hemoglobin (Hb) to detect and classify task conditions in older adults to assess their cognitive capabilities during single and dual task locomotion. We develop an extensive framework for data collection, pre-processing, feature engineering and deep learning and leverage the outstanding learning capabilities of deep neural networks models which surpasses traditional machine learning models by at least 10\% in terms of classification accuracy. To the best of our knowledge, this is the first study to introduce deep learning methods in fNIRS-based single and dual task walking classification in older adults.
\label{sec:c}

\printbibliography
\end{document}